\documentclass[11pt]{article}

\usepackage[a4paper,margin=1in]{geometry}
\usepackage{amsmath,amssymb}
\usepackage{graphicx}
\usepackage{booktabs}
\usepackage{url}
\usepackage{hyperref}
\usepackage{subcaption}

\title{Educational Cone Model in Embedding Vector Spaces}
\author{Yo Ehara\\
Faculty of Education, Tokyo Gakugei University, Japan\\
\texttt{ehara@u-gakugei.ac.jp}}
\date{} 

\begin{document}
\maketitle

\begin{abstract}
Human-annotated datasets with explicit difficulty ratings are essential in 
intelligent educational systems. Although embedding vector spaces are widely used to 
represent semantic closeness and are promising for analyzing text difficulty, the 
abundance of embedding methods creates a challenge in selecting the most suitable 
method. This study proposes the Educational Cone Model, which is a geometric 
framework based on the assumption that easier texts are less diverse (focusing on 
fundamental concepts), whereas harder texts are more diverse. This assumption leads 
to a cone-shaped distribution in the embedding space regardless of the embedding 
method used. The model frames the evaluation of embeddings as an optimization 
problem with the aim of detecting structured difficulty-based patterns. By designing 
specific loss functions, efficient closed-form solutions are derived that avoid costly 
computation. Empirical tests on real-world datasets validated the model’s effectiveness 
and speed in identifying the embedding spaces that are best aligned with difficulty-%
annotated educational texts.
\end{abstract}


\section{Introduction}

Datasets that are annotated with difficulty levels by educators (``difficulty-annotated 
educational datasets'') are essential for developing educational support systems%
~\cite{Arase2022,Hendrycks2020}. Although embedding spaces are widely used to represent 
semantic similarity and are promising for analyzing such datasets, the abundance of 
embedding methods~\cite{Muennighoff2022} poses a challenge in selecting suitable 
methods.

To address this, we propose the Educational Cone Model, which assumes that easier 
items covering fundamental concepts exhibit lower diversity, while more difficult items are 
more diverse. This results in a cone-like structure in the embedding space, independent of 
specific methods. This intuition aligns with the findings of vocabulary acquisition (e.g., Zipf’s 
law), Piaget’s developmental stages~\cite{Piaget1952}, and Bloom’s taxonomy.

We mathematically show that evaluating alignment with this model is reduced to solving an 
optimization problem that identifies a ``difficulty direction'' in the embedding space. By 
designing appropriate loss functions, we derive closed-form solutions, avoiding 
computationally expensive operations such as centroid comparisons.

Empirical evaluations with recent sentence embeddings confirm that the proposed 
model enables efficient selection of embedding methods that are well-aligned with difficulty-%
annotated datasets. 

The contributions of this study are summarized as follows:
\begin{enumerate}
  \item We propose a geometric model that reflects the assumption that easier items are less 
  diverse in the embedding space.
  \item We show that the model can identify a difficulty direction via optimization.
  \item We formulate this as a closed-form optimization problem.
  \item We demonstrate that this solution requires only mean vector differences between difficulty levels.
  \item We empirically validate the effectiveness of the model on real datasets using recent sentence embeddings.
\end{enumerate}

\section{Formulation}

\subsection{Notation}

Let $\{\mathbf{x}_1,\dots,\mathbf{x}_N\}$ denote a set of $N$ embedding vectors, where $\mathbf{x}_i$ is a 
$D$-dimensional vector. We assume that all embedding vectors are normalized; that is, 
$\lVert \mathbf{x} \rVert = 1$, where $\lVert \cdot \rVert$ denotes the Euclidean norm of a vector. 
The proposed method can be applied to both word and sentence embeddings if these 
conditions are satisfied. For simplicity, we refer to both words and sentences as \emph{items} 
throughout this paper. We introduce a $D$-dimensional vector 
$\mathbf{w} \in \mathbb{R}^D$ to represent a direction in the embedding space, with the aim of 
determining the coordinates of $\mathbf{w}$.

\subsection{Educational Cone Model and Difficulty Direction Search Problem}

We first consider the Educational Cone Model and show that, under simple assumptions, it 
aligns with the problem of searching for the difficulty direction in the embedding space. The 
Educational Cone Model assumes that simpler items exhibit lower diversity and more difficult 
items exhibit higher diversity. We interpret the magnitude of diversity in terms of the spatial 
spread within the embedding space. Assuming that simpler items exhibit lower diversity, their
embedding vectors have a smaller spread in the embedding space.

By further refining the notion that simpler items exhibit lower diversity, we assume the 
existence of the ``simplest item.'' Embedding vector spaces are typically structured such that 
semantically similar items are positioned closer together. Hence, in the embedding vector 
space, the simplest item is assumed to reside in the least-spread-out region, represented by a 
point $\mathbf{e}$. If we consider difficulty as a component of ``semantic similarity,'' items closer to $\mathbf{e}$ 
should be simpler, whereas those farther away should be more difficult.

In the Educational Cone Model, suppose that $\mathbf{x}_i$ is simpler than $\mathbf{x}_j$. Based on the above 
discussion, $\mathbf{x}_i$ is closer to the simplest item $\mathbf{e}$ than $\mathbf{x}_j$. Measuring the distance using 
the Euclidean distance and assuming that all $\mathbf{x}$ are normalized to $\lVert \mathbf{x} \rVert = 1$, we 
obtain the following transformations:
\[
  \lVert \mathbf{x}_i - \mathbf{e} \rVert < \lVert \mathbf{x}_j - \mathbf{e} \rVert 
  \;\;\Longleftrightarrow\;\;
  \mathbf{w}^\top \mathbf{x}_i < \mathbf{w}^\top \mathbf{x}_j,
\]
where we define $\mathbf{w} = -\mathbf{e}$. The vector $\mathbf{w}$ represents a direction in the embedding vector 
space. The expression $\mathbf{w}^\top \mathbf{x}_i$ implies that items can be arranged in order along this 
direction. Consequently, the direction $\mathbf{w}$ indicates that moving in this direction within the 
embedding vector space corresponds to increasing difficulty.

\begin{figure}[t]
  \centering
  \begin{subfigure}[t]{0.48\linewidth}
    \centering
    \includegraphics[width=\linewidth]{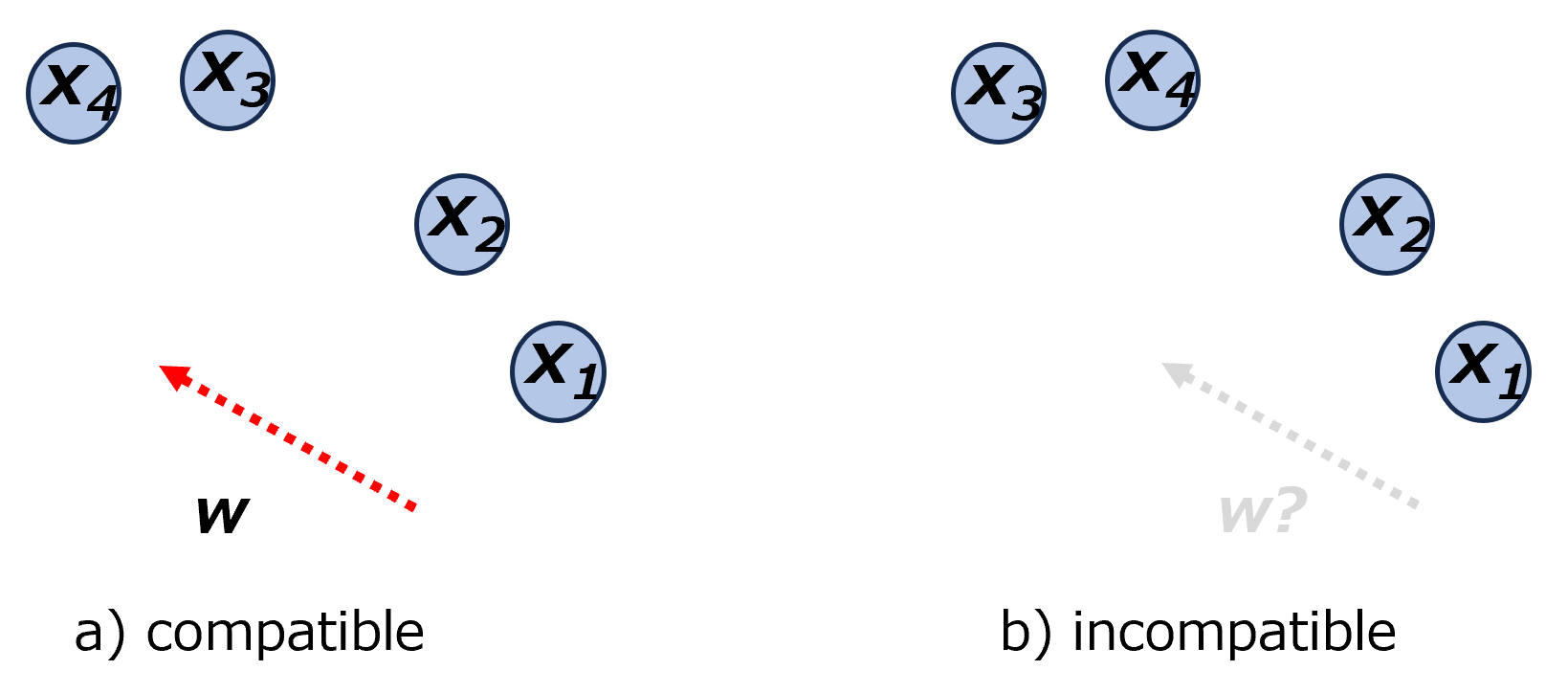}
  \end{subfigure}
  \hfill
  \begin{subfigure}[t]{0.48\linewidth}
    \centering
    \includegraphics[width=\linewidth]{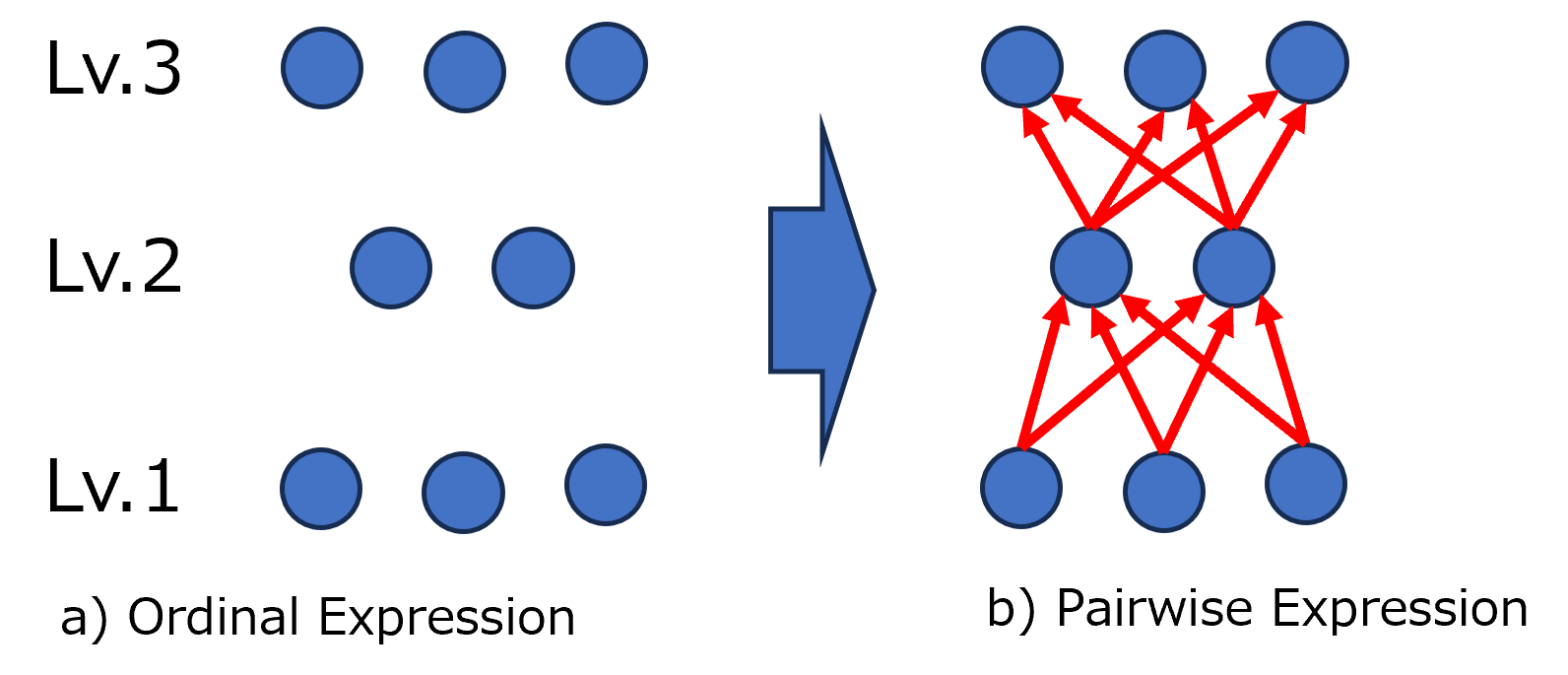}
  \end{subfigure}

  \caption{
  Left: Overview of the proposed method. (a) Considering $\mathbf{x}_1,\mathbf{x}_2,\mathbf{x}_3,\mathbf{x}_4$ as 
  two-dimensional (2D) word/sentence embeddings. $\mathbf{x}_1$ is annotated as simpler than 
  $\mathbf{x}_2$, which in turn is simpler than $\mathbf{x}_3$, etc. If listing points along direction $\mathbf{w}$ aligns 
  with the annotation, the embedding vector set is defined as \emph{compatible} with the 
  annotation. (b) In this case, no direction in the 2D space orders $\mathbf{x}_1,\mathbf{x}_2,\mathbf{x}_3,\mathbf{x}_4$ in the 
  annotated order, so the embedding is defined as \emph{incompatible}. 
  Right: Conversion 
  of difficulty annotations into pairwise constraints.
  }
  \label{fig:overview}
\end{figure}

We provide an intuitive interpretation of the difficulty direction $\mathbf{w}$. Simply put, the difficulty 
direction $\mathbf{w}$ represents the direction in which all items in the dataset (annotation set) appear,
arranged in order of difficulty. Although determining an ideal direction is preferable, it is often 
unrealistic. To address this, we allow slight deviations in the order of difficulty. To this end, we 
first introduce the concept of \emph{compatibility} (Figure~\ref{fig:overview}, left).

In practice, difficulty annotations are often provided in ordinal levels rather than as direct 
pairwise relationships. For example, a question might be annotated as high-school or 
university level rather than being directly compared to another question 
(Figure~\ref{fig:overview}, right). As shown in panel (a), eight items are annotated using three 
levels: Levels~1, 2, and 3. These levels, abbreviated as ``Lv,'' indicate increasing difficulty with 
higher numbers. Levels~1, 2, and 3 have three, two, and three items, respectively. The entire 
ordinal structure can be converted into a directed graph as shown in panel (b). The directed 
edges signify pairwise relationships, and an edge from node $i$ to node $j$ indicates that 
$i$ is easier than $j$. In this manner, without loss of generality, any finite set of ordinal 
annotations can be converted into a mathematically equivalent directed graph.

As illustrated in Figure~\ref{fig:overview}, the difficulty annotations can be converted into a set of 
pairwise constraints. To ensure generality, we define a set of pairwise comparison constraints, 
in which each constraint indicates that one embedding vector is annotated more easily than 
the other. We define the set of pairwise order constraints as
\[
  C = \{(i_1,j_1),\dots,(i_K,j_K)\},
\]
where $K$ denotes the number of constraints. The $k$-th pair $(i_k,j_k)$ represents a single 
constraint.

Here, $i_k \in \{1,\dots,N\}$ and $j_k \in \{1,\dots,N\}$ denote indices of $N$ embedding 
vectors, each of which represents the meaning of an item. The annotation $(i_k,j_k)$ indicates 
that $\mathbf{x}_{i_k}$ is easier than $\mathbf{x}_{j_k}$. For simplicity, we omit the subscript $k$ and denote the 
easier vector as $\mathbf{x}_i$ and the more difficult vector as $\mathbf{x}_j$. We then project these vectors 
onto the direction defined by $\mathbf{w}$ to model the ordering.

Let $\theta_i$ and $\theta_j$ denote the angle between $\mathbf{x}_i$ and the direction $\mathbf{w}$ and the 
angle between $\mathbf{x}_j$ and $\mathbf{w}$, respectively. Our goal is to adjust $\mathbf{w}$ such that the pairwise 
constraints are satisfied:
\[
  \lVert \mathbf{x}_i \rVert \cos\theta_i 
  < \lVert \mathbf{x}_j \rVert \cos\theta_j
  \;\;\Longleftrightarrow\;\;
  \lVert \mathbf{w} \rVert \lVert \mathbf{x}_i \rVert \cos\theta_i 
  < \lVert \mathbf{w} \rVert \lVert \mathbf{x}_j \rVert \cos\theta_j
  \;\;\Longleftrightarrow\;\;
  \mathbf{w}^\top \mathbf{x}_i < \mathbf{w}^\top \mathbf{x}_j
  \;\;\Longleftrightarrow\;\;
  \mathbf{w}^\top (\mathbf{x}_i - \mathbf{x}_j) < 0.
\]

If $K$ pairwise constraints exist, they must all hold simultaneously. However, for practical 
datasets with many items, there may not exist a $\mathbf{w}$ that meets all $K$ constraints. To this 
end, we introduce a slack variable $\xi_k$ to rewrite the inequality constraint into an equality 
constraint:
\[
  \mathbf{w}^\top (\mathbf{x}_{i_k} - \mathbf{x}_{j_k}) + \xi_k = 0.
\]
Intuitively, $\xi_k$ represents the degree to which the constraint is maintained. A larger 
value of $\xi_k$ means that a greater margin is preserved, whereas a smaller value indicates a 
minimal margin enforcement. Although we originally have $\xi_k \ge 0$, we drop this 
constraint to relax the problem and obtain a closed-form solution.

We then consider the following optimization problem:
\begin{equation}
\label{eq:main-opt}
  \max_{\boldsymbol{\xi},\,\mathbf{w}}
  \sum_{k=1}^{K} \xi_k
  \quad\text{s.t.}\quad
  \forall k \in \{1,\dots,K\},\;
  \mathbf{w}^\top (\mathbf{x}_{i_k} - \mathbf{x}_{j_k}) + \xi_k = 0,\;
  \lVert \mathbf{w} \rVert^2 = 1.
\end{equation}

In~\eqref{eq:main-opt}, we impose the norm constraint $\lVert \mathbf{w} \rVert^2 = 1$ to 
obtain a fixed-form solution. Noting that $\xi_k = -\mathbf{w}^\top(\mathbf{x}_{i_k}-\mathbf{x}_{j_k})$ and 
introducing a Lagrange multiplier $\lambda$ to enforce the equality constraint 
$1 - \lVert \mathbf{w} \rVert^2 = 0$, we obtain the following unconstrained optimization 
problem:
\begin{equation}
\label{eq:lagrangian}
  \max_{\mathbf{w}}
  \sum_{k=1}^{K} \bigl[ -\mathbf{w}^\top (\mathbf{x}_{i_k} - \mathbf{x}_{j_k}) \bigr]
  + \lambda\bigl(1 - \lVert \mathbf{w} \rVert^2\bigr).
\end{equation}

Differentiating~\eqref{eq:lagrangian} with respect to $\mathbf{w}$ yields
\[
  \sum_{k=1}^{K} - (\mathbf{x}_{i_k} - \mathbf{x}_{j_k}) - 2\lambda \mathbf{w} = 0,
\]
from which we obtain
\[
  \mathbf{w} \propto \sum_{k=1}^{K} (\mathbf{x}_{j_k} - \mathbf{x}_{i_k}).
\]
This confirms that the optimal direction is simply the mean of the difference vectors, 
normalized to unit length. This property enables us to determine the optimal vector efficiently 
without solving the optimization problem numerically each time.

\section{Experiments}

\subsection{Consistency Experiment Using Word Embeddings and Fine-Grained Annotations}

Consistency experiments were conducted using word embeddings. The dataset used was the
CEFR-J Vocabulary Profile, which contains manually annotated word difficulty levels based 
on the Common European Framework of Reference for Languages (CEFR)%
\footnote{\url{https://github.com/openlanguageprofiles/olp-en-cefrj?tab=readme-ov-file}}.
FastText~\cite{Bhattacharjee2018} was used as the word embedding model. We employed the 
Support Vector Machine (SVM) as the baseline method. The regularization parameter $C$ was 
tuned using validation data, and the optimal value was selected from 
$\{0.1, 1.0, 10.0\}$.

To verify the consistency, we followed a procedure that categorizes words into four levels of 
difficulty. A total of 100 words were randomly selected from the second-easiest category. For 
each word, pairwise constraints were formulated by treating pairs in the easiest category as 
training data and pairs in the third-easiest category as test data. Thus, the proposed convex 
optimization problem was solved. Both the proposed method and SVM achieved 100\% 
accuracy.

In this approach, each word can be regarded as containing 100 different subset
datasets. Another dataset, SVL, provides a more fine-grained 12-level annotation of word 
difficulty%
\footnote{\url{https://eow.alc.co.jp/svl_level12.html}}. Using these 100 datasets, we assessed the 
annotation consistency based on the objective function value of the proposed convex 
optimization problem. As the annotation consistency is measured with respect to the simplest 
word category, words with higher difficulty levels in the 12-level SVL dataset were expected 
to exhibit greater consistency. Spearman’s rank correlation analysis confirmed this 
expectation and yielded a statistically significant correlation ($p < 0.01$). These results 
demonstrate that the proposed method effectively detects consistency in word annotations.

\begin{table}[t]
  \centering
  \caption{Compatibility scores calculated using training data. ``Model'' represents each 
  individual model, and each column corresponds to a pair of CEFR levels. The compatibility 
  score indicates the degree of fit between the level annotations and each embedding. The first 
  two models are 384-dimensional and the latter two models are 1,024-dimensional.}
  \label{tab:compat}
  \begin{tabular}{lcccccc}
    \toprule
    Model & (A1,A2) & (A1,B1) & (A1,B2) & (A2,B1) & (A2,B2) & (B1,B2) \\
    \midrule
    all-MiniLM-L6-v2    & 0.3227 & 0.5741 & 0.9308 & 0.4183 & 0.8599 & 0.7986 \\
    multilingual-e5-small & 0.0835 & 0.1456 & 0.2301 & 0.0835 & 0.1961 & 0.1782 \\
    bge-m3              & 0.1354 & 0.2317 & 0.3930 & 0.1410 & 0.3516 & 0.3336 \\
    multilingual-e5-large & 0.0826 & 0.1435 & 0.2256 & 0.0844 & 0.1952 & 0.1756 \\
    \bottomrule
  \end{tabular}
\end{table}

\subsection{Consistency Experiment Using Sentence Embeddings}

The proposed method is applicable to both word and sentence embeddings. To examine its 
utility further, we evaluated its ability to predict annotation inconsistencies in sentence 
embeddings. For this experiment, we used the CEFR-SP dataset~\cite{Arase2022}, which 
consists of English sentences annotated with four difficulty levels by two annotators. 
Following the same approach as in the word-level experiment, we applied the proposed convex 
optimization problem to the dataset of one annotator. Sentence embeddings were generated 
using \texttt{multilingual-e5-small} \footnote{\url{https://huggingface.co/intfloat/multilingual-e5-small}}.

Owing to the norm constraint in the proposed convex optimization, the solution vector lengths 
were approximately~1.0, enabling a direct comparison of the objective function values across 
sentences. As the value of $\xi$ can be interpreted as a margin, a higher value indicates greater 
consistency. Next, we compared the results with those of the second annotator and calculated 
the Spearman’s rank correlation coefficient between the annotation agreement and objective 
function value. The correlation coefficient was $0.262$, indicating a weak correlation but 
without statistical significance. This suggests that although the proposed method captures 
annotation inconsistencies to a certain extent, further investigation is required.

Using the CEFR-SP dataset, we conducted experiments to identify the most compatible 
sentence embedding model. Each difficulty level pair was treated as a separate dataset and 
the compatibility score was computed using the proposed method. Table~\ref{tab:compat} 
presents the compatibility scores obtained from the sentence embedding experiments. Each 
column corresponds to a CEFR difficulty level pair, and the table summarizes the embedding 
models used. Higher scores indicate a better fit between the embedding space and the 
annotations.

\subsection{Prediction Experiment Using SVM}

Finally, we conducted a prediction experiment using SVM with sentence embeddings. We 
utilized (A1,B1) as the training datasets and (B1,B2) as the test datasets. The four embedding 
models summarized in Table~\ref{tab:compat} achieved prediction accuracies of 
$0.26$, $0.23$, $0.63$, and $0.37$, respectively. Notably, the first two embeddings had 
384 dimensions, whereas the latter two had 1{,}024 dimensions. We observed that when the 
dimensionality was the same, embeddings with higher compatibility scores tended to yield 
superior predictive performance.

This finding suggests that, given the same dimensionality, the compatibility score can reliably 
estimate the predictive performance of SVM without requiring access to test data. 
Consequently, our approach enables an efficient assessment of embedding quality for text 
difficulty prediction, without the need for computationally expensive model training on each 
embedding.

\section{Related Work}

As an early study on the relationship between contextualized embedding vectors and 
difficulty, Ehara~\cite{Ehara2022} expressed the difficulty of word examples by the frequency 
of nearby examples in the embedding vector space. Recently, Ehara~\cite{Ehara2025} 
proposed a method for controlling the difficulty of educational items by combining linear 
interpolation of sentence embeddings with different difficulty levels in the embedding space 
and a technique called Inverse Embedding, which generates sentences from embedding 
vector coordinates.

Several recent studies have investigated the interpretability of embedding vectors%
~\cite{Vasilyev2024,LiLi2024,Chen2024}. Vasilyev et al.~\cite{Vasilyev2024} examined 
multilingual embeddings using linear transformations under the assumption that embeddings 
are orthogonally aligned across languages. Li and Li~\cite{LiLi2024} analyzed relationships 
between contextualized embeddings from large-scale language models and static embeddings 
such as sentence embeddings. Chen et al.~\cite{Chen2024} proposed methods for 
constructing finer-grained embeddings for thematic sentence representations. Building on this 
research, our study focuses on the consistency of ordinal annotations and presents a method 
for quantifying this consistency.

\section{Conclusion}

This paper has introduced the Educational Cone Model, which posits that easier items 
exhibit lower diversity than more difficult items in embedding vector spaces owing to their 
inherent complexity. This principle aligns with findings of previous studies on educational 
classifications. Leveraging this insight, we proposed a computationally efficient method for 
determining the directional orientation of educational items within an embedding space.

As a convex optimization approach, our method guarantees convergence to a global optimum, 
making it robust to initial conditions and free from approximation errors. This contrasts 
sharply with recent neural-network-based methods, which are typically sensitive to 
initialization and only achieve locally optimal solutions.

The empirical evaluation demonstrated a statistically significant correlation between our 
method and fine-grained annotation consistency in word difficulty datasets. We further 
conducted experiments using the proposed method with recent sentence embeddings. The 
proposed objective function accurately predicted the performance of embeddings as 
predictive features without using test data. This result substantiates our method’s ability to 
capture the ``cone'' structure of learning material difficulty within an embedding space 
effectively.

Future research directions include extending the approach to a broader range of educational 
tasks, including STEM-related questions. Notably, the interpretability of our method allows the 
identification of subject-specific difficulty dimensions, such as those unique to physics or 
chemistry, by treating difficulty as a directional property within the embedding space.

\section*{Acknowledgements}

This work was supported by JSPS KAKENHI Grant Number 22K12287 and by JST, PRESTO 
Grant Number JPMJPR2363.

\end{document}